\documentclass[runningheads]{llncs}
\usepackage{times}
\usepackage{inconsolata}
\usepackage{latexsym}

\usepackage{microtype}

\usepackage{hyperref}

\usepackage{microtype}
\usepackage{graphicx}
\usepackage{subfigure}
\usepackage{booktabs} %
\usepackage{multirow}

\usepackage{xspace}
\usepackage{enumitem}
\usepackage{color, colortbl}
\usepackage[normalem]{ulem}

\usepackage{graphicx}

\usepackage{glossaries}
\newacronym{kg}{KG}{knowledge graph}
\newacronym{lp}{LP}{Link prediction}

\glsdisablehyper

\usepackage{tikz}
\usetikzlibrary{fit}

\usepackage{amsmath}
\usepackage{amssymb}

\makeatletter
\newcommand{\printfnsymbol}[1]{%
  \textsuperscript{\@fnsymbol{#1}}%
}
\makeatother

\title{Improving Inductive Link Prediction Using Hyper-Relational Facts}

\author{Mehdi Ali\inst{1,2}\thanks{equal contribution} \and
Max Berrendorf\inst{3}\printfnsymbol{1} \and
Mikhail Galkin \inst{4} \and
Veronika Thost \inst{5} \and
Tengfei Ma \inst{5} \and
Volker Tresp \inst{3,6} \and
Jens Lehmann\inst{1,2}
}
\authorrunning{Ali \textit{et al.}}

\institute{Smart Data Analytics Group, University of Bonn, Germany \\
\email{\{mehdi.ali,jens.lehmann\}@cs.uni-bonn.de} \and
Fraunhofer Institute for Intelligent Analysis and
Information Systems (IAIS), Sankt Augustin and Dresden, Germany\\
\email{\{mehdi.ali,jens.lehmann\}@iais.fraunhofer.de} \and
Ludwig-Maximilians-Universit\"at M\"unchen, Munich, Germany \\
\email{\{berrendorf,tresp\}@dbs.ifi.lmu.de} \and
Mila, McGill University \\
\email{mikhail.galkin@mila.quebec}
\and
IBM Research, MIT-IBM Watson AI Lab\\
\email{vth@zurich.ibm.com}, \email{tengfei.ma1@ibm.com}
\and 
Siemens AG, Munich, Germany \\
\email{volker.tresp@siemens.com} 
}

\begin{document}

\newcommand{\rel}{\mathcal{R}\xspace}
\newcommand{\en}{\mathcal{E}\xspace}
\newcommand{\eh}{\mathcal{E}_H\xspace}
\newcommand{\et}{\mathcal{E}_T\xspace}
\newcommand{\equ}{\mathcal{E}_Q\xspace}
\newcommand{\es}{\mathcal{E}_{\bullet}\xspace}
\newcommand{\eu}{\mathcal{E}_{\circ}\xspace}

\newcommand{\strain}{\mathcal{S}_{train}\xspace}
\newcommand{\seval}{\mathcal{S}_{eval}\xspace}
\newcommand{\sinf}{\mathcal{S}_{\mathit{inf}}\xspace}

\newcommand{\pset}{\mathfrak{P}\xspace}

 \maketitle

\begin{abstract}
For many years, link prediction on \glspl{kg} has been a purely transductive task, not allowing for reasoning on unseen entities. 
Recently, increasing efforts are put into exploring semi- and fully inductive scenarios, enabling inference over unseen and emerging entities.
Still, all these approaches only consider triple-based \glspl{kg}, whereas their richer counterparts, hyper-relational \glspl{kg} (e.g., Wikidata), have not yet been properly studied.
In this work, we classify different inductive settings and study the benefits of employing hyper-relational \glspl{kg} on a wide range of semi- and fully inductive link prediction tasks powered by recent advancements in graph neural networks. 
Our experiments on a novel set of benchmarks show that qualifiers over typed edges can lead to performance improvements of 6\% of absolute gains (for the Hits@10 metric) compared to triple-only baselines. 
Our code is available at \url{https://github.com/mali-git/hyper_relational_ilp}.
\end{abstract}

\section{Introduction}

\begin{figure}[t]
    \centering
    \includegraphics[width=0.68\textwidth]{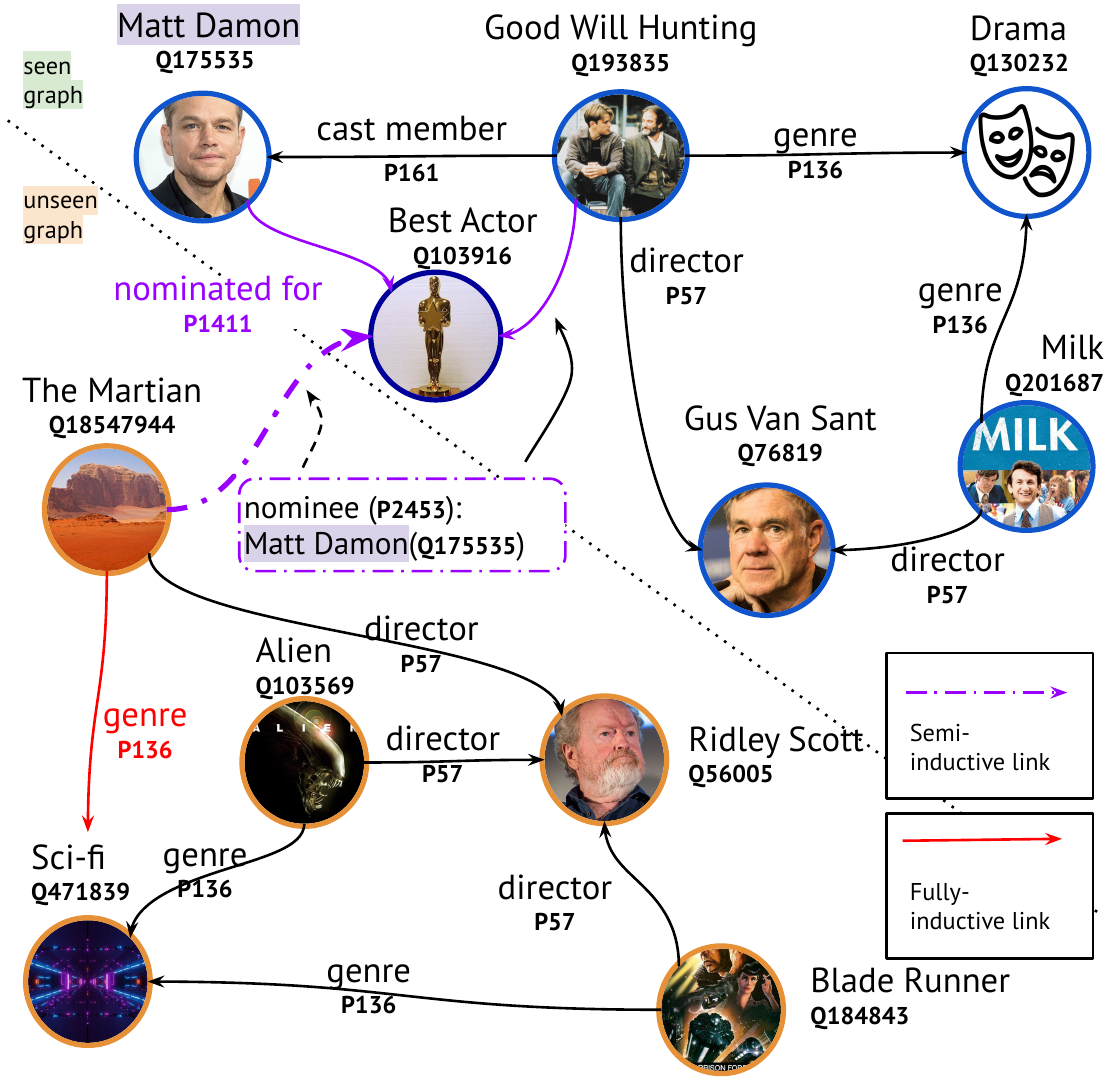}
    \caption{
    Different types of inductive LP. 
    Semi-inductive: the link between \texttt{The Martian} and \texttt{Best Actor} from the seen graph. 
    Fully-inductive: the \texttt{genre} link between unseen entities given a new unseen subgraph \emph{at inference time}. 
    The qualifier \texttt{(nominee: Matt Damon)} over the original relation \texttt{nominated for} allows to better predict the semi-inductive link.
    }
    \label{fig:settings}
\end{figure}

Knowledge graphs are notorious  for their sparsity and incompleteness~\cite{DBLP:conf/kdd/0001GHHLMSSZ14}, so that predicting missing links has been one of the first applications of machine learning and embedding-based methods over KGs~\cite{DBLP:conf/icml/NickelTK11,DBLP:conf/nips/BordesUGWY13}.
A flurry~\cite{DBLP:journals/corr/abs-2006-13365,DBLP:journals/corr/abs-2002-00388} of such algorithms has been developed over the years, and most of them share certain commonalities, i.e., they operate over \emph{triple-based} KGs in the \emph{transductive} setup, where all entities are known at training time. 
Such approaches can neither operate on unseen entities, which might emerge after updating the graph, nor on new (sub-)graphs comprised of completely new entities.
Those scenarios are often unified under the  \emph{inductive} link prediction (LP) setup.
A variety of NLP tasks building upon KGs have inductive nature, for instance, entity linking or information extraction.
Hence, being able to work in inductive settings becomes crucial for KG representation learning algorithms. 
For instance (cf. Fig.~\ref{fig:settings}), the  \texttt{director}-\texttt{genre} pattern from the seen graph allows to predict a missing \texttt{genre} link for \texttt{The Martian} in the unseen subgraph.

Several recent approaches~\cite{DBLP:conf/icml/TeruDH20,daza2020inductive} tackle an inductive LP task, but they usually focus on a specific inductive setting. %
Furthermore, their underlying KG structure is still based on triples.
On the other hand, new, more expressive KGs like Wikidata~\cite{DBLP:journals/cacm/VrandecicK14} exhibit a \emph{hyper-relational} nature where each triple (a typed edge in a graph) can be further instantiated with a set of explicit relation-entity pairs, known as \emph{qualifiers} in the Wikidata model.
Recently, it was shown~\cite{galkin2020message} that employing hyper-relational KGs yields significant gains in the transductive LP task compared to their triple-only counterparts. 
But the effect of such KGs on inductive LP is unclear. Intuitively (Fig.~\ref{fig:settings}), the \texttt{(nominee: Matt Damon)} qualifier provides a helpful signal to predict \texttt{Best Actor} as an object of \texttt{nominated for} of \texttt{The Martian} given that \texttt{Good Will Hunting} received such an award with the same nominee.

In this work, we systematically study hyper-relational KGs in different inductive settings:
\begin{itemize}[leftmargin=*,noitemsep,topsep=0pt] 
\item We propose a classification of inductive LP scenarios that describes the settings formally and, to the best of our knowledge, integrates all relevant existing works. Specifically, we distinguish \emph{fully-inductive} scenarios, where target links are to be predicted in a new subgraph of unseen entities, and \emph{semi-inductive} ones where unseen nodes have to be connected to a known graph.
\item We then adapt two existing baseline models for the two inductive LP tasks probing them in the hyper-relational settings.
 \item Our experiments suggest that models supporting hyper-relational facts indeed improve link prediction in both %
 inductive settings compared to strong triple-only baselines by more than 6\% Hits@10.
\end{itemize}

\section{Background}
We assume the reader to be familiar with the standard link prediction setting (e.g. from~\cite{DBLP:conf/icml/NickelTK11}) and introduce the specifics of the setting with qualifiers.

\subsection{Statements: Triples plus Qualifiers}
Let $G = (\mathcal{E}, \mathcal{R}, \mathcal{S})$ be a hyper-relational KG where $\mathcal{E}$ is a set of entities, $\mathcal{R}$ is a set of relations, and $\mathcal{S}$ a set of statements.
Each statement can be formalized as a 4-tuple $(h, r, t, q)$ of a head and tail entity\footnote{We use \emph{entity} and \emph{node} interchangeably} $h, t \in \mathcal{E}$, a relation $r \in \mathcal{R}$, and a set of qualifiers, which are relation-entity pairs $q \subseteq \pset (\mathcal{R} \times \mathcal{E})$ where $\pset$ denotes the power set.
For example, Fig.~\ref{fig:settings} contains a statement \texttt{(Good Will Hunting, nominated for, Best Actor, \{(nominee, Matt Damon)\})} where \texttt{(nominee, Matt Damon)} is a qualifier pair for the main triple.
We define the set of all possible statements as set
\[
\mathbb{S}(\eh, \rel, \et, \equ) = \eh\times \rel \times \et \times \pset(\rel \times \equ)
\]
with a set of relations $\rel$, a set of head, tail and qualifier entities $\eh, \et, 
\equ \subseteq \en$. %
Further, $\strain$ is the set of training statements and %
$\seval$ are evaluation statements.
We assume that we have a feature vector %
$\mathbf{x}_e \in \mathbb{R}^d$ associated with each entity $e \in \mathcal{E}$.
Such feature vectors can, for instance, be obtained from entity descriptions available in some KGs or represent topological features such as Laplacian eigenvectors~\cite{DBLP:conf/nips/BelkinN01} or regular graph substructures~\cite{DBLP:journals/corr/abs-2006-09252}.
In this work, we focus on the setting with one fixed set of known relations. 
That is, we do not require $\mathbf{x}_r \in \mathbb{R}^d$ features for relations and rather learn relation embeddings during training. %

\subsection{Expressiveness} %

Models making use of qualifiers are strictly more expressive than those which do not:
Consider the following example with two statements, $s_1 = (h, r, t, q_1)$ and $s_2 = (h, r, t, q_2)$, sharing the same triple components, but differing in their qualifiers, such that $s_1|q_1 = False$ and $s_2|q_2 = True$.
For a model $f_{NQ}$ not using qualifiers, i.e., only using the triple component $(h, r, t)$, we have $f_{NQ}(s_1) = f_{NQ}(s_2)$.
In contrast, a model $f_Q$ using qualifiers can predict $f_Q(s_1) \neq f_Q(s_2)$, thus being strictly more expressive.

\section{Inductive Link Prediction}

\begin{table*}[t]
\centering
\resizebox{\textwidth}{!}{
\begin{tabular}{@{}lccccc@{}}
\toprule
Named scenario & %
$\sinf$ & Unseen $\leftrightarrow$ Unseen & Unseen $\leftrightarrow$ Seen
& Scoring against & In our framework \\ \midrule
Out-of-sample~\cite{DBLP:conf/emnlp/AlbooyehGK20}  & $k$-shot & - & \checkmark & $E_{tr}$ & SI \\
Unseen entities~\cite{clouatre2020mlmlm} & $k$-shot & - & \checkmark & $E_{tr}$ & SI \\
Inductive~\cite{DBLP:conf/semweb/BhowmikM20} & $k$-shot & - & \checkmark & $E_{tr}$ & SI \\
Inductive~\cite{DBLP:conf/icml/TeruDH20} & new graph & \checkmark & - & $E_{\mathit{inf}}$ & FI \\
Transfer~\cite{daza2020inductive} & new graph & \checkmark & - & $E_{\mathit{inf}}$ & FI  \\
Dynamic~\cite{daza2020inductive} & $k$-shot + new graph & \checkmark & \checkmark & $E_{tr} \cup E_{\mathit{inf}}$ & FI / SI \\
Out-of-graph~\cite{DBLP:conf/nips/BaekLH20}  & $k$-shot + new graph  & \checkmark & \checkmark & $E_{tr} \cup E_{\mathit{inf}}$ & FI / SI \\
Inductive~\cite{DBLP:journals/corr/abs-2009-09263} & $k$-shot + new graph & \checkmark & \checkmark & $E_{tr} \cup E_{\mathit{inf}}$ & FI / SI \\

 \bottomrule
\end{tabular}}
\caption{Inductive LP in the literature, a discrepancy in terminology. %
The approaches differ in the kind of auxiliary statements $\sinf$ used at inference time: 
in whether they contain entities seen during training $E_{\mathit{tr}}$ and whether new entities $E_{\mathit{inf}}$ are connected to seen ones (\emph{k-shot} scenario), or (only) amongst each other, in a new graph. Note that the evaluation settings also~vary.
}
\label{tab:lp_papers}
\end{table*}

Recent works (cf. Table~\ref{tab:lp_papers})
have pointed out the practical relevance of different inductive LP scenarios.
However, there exists a terminology gap as different authors employ different names for describing conceptually the same task or, conversely, use the same \emph{inductive LP} term for practically different setups.  
We propose a unified framework that provides an overview of the area %
and describes the settings formally. %

Let $\es$ denote the set of entities occurring in the training statements $\strain$ at any position (head, tail, or qualifier), and $\eu \subseteq \en \setminus \es$ denote a set of unseen entities. 
In the \emph{transductive} setting, all entities in the evaluation statements are seen during training, i.e., $\seval \subseteq \mathbb{S}(\es, \rel, \es, \es)$.
In contrast, in \emph{inductive} settings, $\seval$, used in validation and testing, may contain unseen entities. In order to be able to learn representations for these entities at inference time, inductive approaches may consider an additional set $\sinf$ of \emph{inference statements} about (un)seen entities; of course $\sinf\cap\seval=\emptyset$. %

The \emph{fully-inductive} setting (FI) is akin to transfer learning where link prediction is performed over a set of entities not seen before, i.e., $\seval \subseteq \mathbb{S}(\eu, \rel, \eu, \eu)$. This is made possible by providing an auxiliary inference graph $\sinf \subseteq \mathbb{S}(\eu, \rel, \eu, \eu)$ containing statements about the unseen entities in $\seval$.
For instance, in Fig.~\ref{fig:settings}, the training graph is comprised of entities \texttt{Matt Damon, Good Will Hunting, Best Actor, Gus Van Sant, Milk, Drama}. 
The inference graph contains new entities \texttt{The Martian, Alien, Ridley Scott, Blade Runner, Sci-fi} with one missing link to be predicted.
The fully-inductive setting is considered in \cite{DBLP:conf/icml/TeruDH20,daza2020inductive}.

In the \emph{semi-inductive} setting (SI), new, unseen entities are to be connected to seen entities, i.e., $\seval \subseteq \mathbb{S}(\es, \rel, \eu, \es) \cup \mathbb{S}(\eu, \rel, \es, \es)$. %
Illustrating with Fig.~\ref{fig:settings}, \texttt{The Martian} as the only unseen entity connecting to the seen graph, the semi-inductive statement connects \texttt{The Martian} to the seen \texttt{Best Actor}. Note that there are other practically relevant examples beyond KGs, such as predicting interaction links between a new drug and a graph containing existing proteins/drugs \cite{dti-overview,DBLP:journals/corr/abs-2012-05716}.
We hypothesize that, in most scenarios, we are not given any additional information about the new entity, and thus have $\sinf=\emptyset$; we will focus on this case in this paper. However, the
variation where $\sinf$ may contain $k$ statements connecting the unseen entity to seen ones has been considered too \cite{DBLP:conf/emnlp/AlbooyehGK20,DBLP:conf/semweb/BhowmikM20,clouatre2020mlmlm} and is known as \emph{k-shot learning} scenario.

A mix of the fully- and semi-inductive settings 
where evaluation statements may contain two instead of just one unseen entity is studied in~\cite{daza2020inductive,DBLP:conf/nips/BaekLH20,DBLP:journals/corr/abs-2009-09263}.
That is, unseen entities might be connected to the seen graph, i.e., $\seval$ may contain seen entities, and, at the same time, the unseen entities might be connected to each other; i.e, $\sinf\neq\emptyset$.
Our framework is general enough to allow $\seval$ to contain new, unseen relations $r$ having their features $\mathbf{x}_r$ at hand. 
Still, to the best of our knowledge, research so far has focused on the setting where all relations are seen in training; %
we will do so, too. 

We %
hypothesize that qualifiers, being explicit attributes over typed edges, provide a strong inductive bias for LP tasks.
In this work, for simplicity, we require both qualifier relations and entities to be seen in the training graph, i.e., $\mathcal{E}_Q \subseteq \es$ and $\mathcal{R}_Q \subseteq \mathcal{R}$, although the framework accommodates a more general case of unseen qualifiers given their respective features.

\section{Approach}
\label{sec:approach}

Both semi- and fully-inductive tasks assume node features to be given. %
Recall that relation embeddings are learned and, often, to reduce the computational complexity, their dimensionality is smaller than that of node features.

\subsection{Encoders} \label{sec:approach_enc}

In the semi-inductive setting, an unseen entity arrives without any graph structure pointing to existing entities, i.e., $\sinf=\emptyset$.
This fact renders message passing approaches~\cite{DBLP:conf/icml/GilmerSRVD17} less applicable, so we resort to a simple linear layer to project all entity features (including those of qualifiers) into the relation space:
$
    \phi %
    : \mathbb{R}^{d_f} \rightarrow \mathbb{R}^{d_r}
$

In the fully inductive setting, we are given a non-empty inference graph $\sinf\neq\emptyset$, 
and we probe two encoders: (i) the same linear projection of features as in the semi-inductive scenario which does not consider the graph structure; (ii) GNNs which can naturally work in the inductive settings~\cite{DBLP:journals/corr/abs-2005-03675}. 
However, the majority of existing GNN encoders for multi-relational KGs like CompGCN~\cite{DBLP:conf/iclr/VashishthSNT20} are limited to only triple KG representation. 
To the best of our knowledge, only the recently proposed \textsc{StarE}~\cite{galkin2020message} encoder supports hyper-relational KGs which we take as a basis for our inductive model.
Its aggregation formula is:
\begin{equation}
     \mathbf{x}_v' = f \left( \sum_{(u,r) \in \mathcal{N}(v)} \mathbf{W}_{\lambda(r)} \phi_r ( \mathbf{x}_u, \gamma (\mathbf{x}_r,\mathbf{x}_{q} )_{vu} ) \right)
\label{eq:stare}     
\end{equation}
where $\gamma$ %
is a function that infuses the vector of aggregated qualifiers $\mathbf{x}_q$ into the vector of the main relation $\mathbf{x}_r$. 
The output of the GNN contains updated node and relation features based on the adjacency matrix $A$ and qualifiers $Q$:
\begin{align*}
    \mathbf{X',R'}=\textsc{StarE}(A,\mathbf{X},\mathbf{R}, Q)
\end{align*}

Finally, in both inductive settings, we linearize an input statement in a sequence using a padding index where necessary: $[\mathbf{x}_h', \mathbf{x}_r', \mathbf{x}_{q_1^r}', \mathbf{x}_{q_1^e}', [\text{PAD}], \ldots]$.
Note that statements can greatly vary in length depending on the amount of qualifier pairs, and padding mitigates this issue.

\subsection{Decoder}\label{sec:approach_dec}
Given an encoded sequence, we use the same Transformer-based decoder for all settings:
\begin{align*}
f(h, r, t, q)& = g(\mathbf{x}_h', \mathbf{x}_r', \mathbf{x}_{q_1^r}', \mathbf{x}_{q_1^e}', \ldots)^T \mathbf{x}_t'
\ \ \text{with}\\
g(\mathbf{x}_1', \ldots, \mathbf{x}_k)&= \text{Agg}(\text{Transformer}([\mathbf{x}_1', \ldots, \mathbf{x}_k']))
\end{align*}

In this work, we evaluated several aggregation strategies and found a simple mean pooling over all non-padded sequence elements to be preferable.
Interaction functions of the form $f(h,r,t,q) = f_1(h,r,q)^T f_2(t)$ are particularly well-suited for fast 1-N scoring for tail entities, since the first part only needs to be computed only once.

Here and below, we denote the linear encoder + Transformer decoder model as QBLP (that is, Qualifier-aware BLP, an extension of BLP~\cite{daza2020inductive}), and the \textsc{StarE} encoder + Transformer decoder, as \textsc{StarE}.

\subsection{Training}

In order to compare results with triple-only approaches, we train the models, as usual, on the subject and object prediction tasks. %
We use stochastic local closed world assumption (sLCWA) and the local closed world assumption (LCWA) commonly used in the KG embedding literature~\cite{DBLP:journals/corr/abs-2006-13365}. 
Particular details on sLCWA and LCWA are presented in Appendix~\ref{app:training}. 
Importantly, in the semi-inductive setting, the models score against all entities in the training graph $E_{tr}$ in both training and inference stages.
In the fully-inductive scenario, as we are predicting links over an unseen graph, the models score against all entities in $E_{tr}$ during training and against unseen entities in the inference graph $E_{\mathit{inf}}$ during inference.

\section{Datasets}
\label{sec:datasets}

\begin{table*}[t]
\centering
\resizebox{\textwidth}{!}{
\begin{tabular}{@{}cl*{12}{r}@{}}
\toprule
\multirow{2}{*}{Type} & \multirow{2}{*}{Name} & \multicolumn{3}{c}{Train} & \multicolumn{3}{c}{Validation} & \multicolumn{3}{c}{Test} & \multicolumn{3}{c}{Inference}  \\ \cmidrule(r){3-5}  \cmidrule(r){6-8}  \cmidrule(r){9-11}  \cmidrule(r){12-14}  
 &  & $S_{tr}~(Q\%)$ & $E_{tr}$ & $R_{tr}$ & $S_{vl}~(Q\%)$ & $E_{vl}$ & $R_{vl}$ & $S_{ts}~(Q\%)$ & $E_{ts}$ & $R_{ts}$ & $S_{\mathit{inf}} (Q\%)$ & $E_{\mathit{inf}}$ & $R_{\mathit{inf}}$ \\ \midrule
SI & WD20K (25) & 39,819 (\phantom{0}30\%) & 17,014 & 362 & 4,252 (\phantom{0}25\%) & 3544 & 194 & 3,453 (\phantom{0}22\%) & 3028 & 198 & - & - & - \\
SI & WD20K (33) & 25,862 (\phantom{0}37\%) & 9251 & 230 & 2,423 (\phantom{0}31\%) & 1951 & 88 & 2,164 (\phantom{0}28\%) & 1653 & 87 & - & - & - \\
FI & WD20K (66) V1 & 9,020 (\phantom{0}85\%) & 6522 & 179 & 910 (\phantom{0}45\%) & 1516 & 111 & 1,113 (\phantom{0}50\%) & 1796 & 110 & 6,949 (\phantom{0}49\%) & 8313 & 152 \\
FI & WD20K (66) V2 & 4,553 (\phantom{0}65\%) & 4269 & 148 & 1,480 (\phantom{0}66\%) & 2322 & 79 & 1,840 (\phantom{0}65\%) & 2700 & 89 & 8,922 (\phantom{0}58\%) & 9895 & 120 \\
FI & WD20K (100) V1 & 7,785 (100\%) & 5783 & 92 & 295 (100\%) & 643 & 43 & 364 (100\%) & 775 & 43 & 2,667 (100\%) & 4218 & 75 \\
FI & WD20K (100) V2 & 4,146 (100\%) & 3227 & 57 & 538 (100\%) & 973 & 43 & 678 (100\%) & 1212 & 42 & 4,274 (100\%) & 5573  & 54 \\ 
\bottomrule
\end{tabular}}
\caption{Semi-inductive (SI) and fully-inductive (FI) datasets. $S_{ds} (Q\%)$ denotes the number of statements with the qualifiers ratio in train ($ds=tr$), validation ($ds=vl$), test ($ds=ts$), and inductive inference ($ds=\mathit{inf}$) splits. $E_{ds}$ is the number of distinct entities. $R_{ds}$ is the number of distinct relations. $S_{inf}$ is a basic graph for $vl$ and $ts$ in the FI scenario. }
\label{tab:datasets}
\end{table*}

We take the original transductive splits of the WD50K~\cite{galkin2020message} family of hyper-relational datasets as a leakage-free basis for sampling our semi- and fully-inductive datasets which we denote by WD20K.

\subsection{Fully-Inductive Setting}
We start with extracting \emph{statement entities} $\mathcal{E}'$, and sample $n$ entities and their $k$-hop neighbourhood to form the statements $(h,r,t,q)$ of the transductive train graph $S_{train}$.
From the remaining $\mathcal{E}' \setminus \mathcal{E}_{train}$ and $S \setminus S_{train}$ sets we sample $m$ entities with their $l$-hop neighbourhood to form the statements $S_{ind}$ of the inductive graph. 
The entities of $S_{ind}$ are disjoint with those of the transductive train graph. 
Further, we filter out all statements in $S_{ind}$ whose relations (main or qualifier) were not seen in $S_{train}$.
Then, we randomly split $S_{ind}$ with the ratio about 55\%/20\%/25\% into inductive inference, validation, and test statements, respectively.
The evaluated models are trained on the transductive train graph $S_{train}$. 
During inference, the models receive an unseen inductive inference graph from which they have to predict validation and test statements.
Varying $k$ and $l$, we sample two different splits: V1 has a larger training graph with more seen entities whereas V2 has a bigger inductive inference graph.

\subsection{Semi-Inductive Setting}
Starting from all statements, we extract all entities occurring as head or tail entity in any statement, denoted by $\mathcal{E}'$ and named \emph{statement entities}.
Next, we split the set of statement entities into a train, validation and test set: $\mathcal{E}_{train}, \mathcal{E}_{validation}, \mathcal{E}_{test}$.
We then proceed to extract statements $(h, r, t, q) \in S$ with one entity ($h/t$) in $\mathcal{E}_{train}$ and the other entity in the corresponding statement entity split.
We furthermore filter the qualifiers to contain only pairs where the entity is in a set of allowed entities, formed by $\mathcal{A}_{split} = \mathcal{E}_{train} \cup \mathcal{E}_{split}$, with split being train/validation/test.
Finally, since we do not assume relations to have any features, we do not allow unseen relations.
We thus filter out relations which do not occur in the training statements.

\subsection{Overview}
To measure the effect of hyper-relational facts on both inductive LP tasks, we sample several datasets varying the ratio of statements with and without qualifiers. 
In order to obtain the initial node features we mine their English surface forms and descriptions available in Wikidata as \texttt{rdfs:label} and \texttt{schema:description} values. 
The surface forms and descriptions are concatenated into one string and passed through the Sentence BERT~\cite{reimers-2019-sentence-bert} encoder based on RoBERTa~\cite{DBLP:journals/corr/abs-1907-11692} to get 1024-dimensional vectors.
The overall datasets statistics is presented in Table~\ref{tab:datasets}.

\section{Experiments}

We design our experiments to investigate whether the incorporation of qualifiers improves inductive link prediction. In particular, we investigate the fully-inductive setting (Section~\ref{exp_full}) and the semi-inductive setting (Section~\ref{exp_semi}). We analyze the impact of the qualifier ratio (i.e., the number of statements with qualifiers) and the dataset's size on a model's performance.

\subsection{Experimental Setup}

\begin{table*}[t]
\centering
\resizebox{\textwidth}{!}{
\begin{tabular}{lccccccccccc}
\toprule
\multirow{2}{*}{Model} & \multirow{2}{*}{\#QP} & \multicolumn{5}{c}{WD20K (100) V1} & \multicolumn{5}{c}{WD20K (100) V2} \\ \cmidrule(r){3-7}  \cmidrule(r){8-12}  
& & AMR(\%) &  MRR(\%) &  H@1(\%) &  H@5(\%) &  H@10(\%) & AMR(\%) &  MRR(\%) &  H@1(\%) &  H@5(\%) &  H@10(\%) \\
\midrule
  BLP &   0 &     22.78 &    5.73 &       1.92 &       8.22 &       12.33 &  36.71 &    3.99 &       1.47 &       4.87 &        9.22 \\
  CompGCN &  0  &   37.02 &   10.42 &       \underline{5.75} &      15.07 &       18.36 & 74.00 &    2.55 &       0.74 &       3.39 &        5.31 \\
  QBLP   &   0 &      28.91 &    5.52 &       1.51 &       8.08 &       12.60 & 35.38 &    4.94 &       2.58 &       5.46 &        9.66 \\
\midrule
StarE  &   2 &   41.89 &    9.68 &       3.73 &      \textbf{16.57} &       20.99 & 40.60 &    2.43 &       0.45 &       3.86 &        6.17 \\
StarE  &   4 &      35.33 &   10.41 &       4.82 &      15.84 &       21.76 & 37.16 &    5.12 &       1.41 &      7.93 &       12.89 \\
StarE & 6 &   34.86 &   \textbf{11.27} &       \textbf{6.18} &      15.93 &       21.29 & 47.35 &    4.99 &       1.92 &       6.71 &       11.06 \\
  QBLP  &   2 &   \textbf{18.91} &   10.45 &       3.73 &      16.02 &       \underline{22.65} &  \textbf{28.03} &    \textbf{6.69} &       \textbf{3.49} &       \underline{8.47} &       12.04 \\
  QBLP   &   4 &   \underline{20.19} &   \underline{10.70} &       3.99 &      \underline{16.12} &       \textbf{24.52}&   \underline{31.30} &    5.87 &       2.37  &       7.85 &       \textbf{13.93} \\
  QBLP  &   6 &     23.65 &    7.87 &       2.75 &             10.44 &       17.86  &   34.35 &    \underline{6.53} &       \underline{2.95} &       \textbf{9.29} &       \underline{13.13} \\
\bottomrule
\end{tabular}}
\caption{Results on FI WD20K (100) V1 \& V2. \#QP denotes the number of qualifier pairs used in each statement (including padded pairs). Best results \textbf{in bold}, second best \underline{underlined}.}
\label{full_wd50_100_inductive}
\end{table*}

\begin{table*}[t]
\centering
\resizebox{\textwidth}{!}{
\begin{tabular}{lccccccccccc}
\toprule
   \multirow{2}{*}{Model} & \multirow{2}{*}{\#QP} & \multicolumn{5}{c}{WD20K (66) V1} & \multicolumn{5}{c}{WD20K (66) V2} \\ \cmidrule(r){3-7}  \cmidrule(r){8-12}  
& & AMR(\%) &  MRR(\%) &  H@1(\%) &  H@5(\%) &  H@10(\%) & AMR(\%) &  MRR(\%) &  H@1(\%) &  H@5(\%) &  H@10(\%) \\
\midrule
    BLP  &   0 &   34.96 &    2.10 &       0.45 &         2.29 &        4.44 & 45.29 &    1.56 &       0.27 &         1.88 &        3.35 \\
    CompGCN &   0 &   35.99 &    5.80 &       2.38 &        8.93 &       12.79 & 47.24 &    \underline{2.56} &       \underline{1.17} &       3.07 &        4.46 \\
    QBLP &   0 &   35.30 &    3.69 &       1.30 &       4.85 &        7.14 & 42.48 &    0.94 &       0.08 &       0.79 &        1.82 \\
\midrule
    StarE  &   2 &   37.72 &    \textbf{6.84} &       \textbf{3.24} &         \textbf{9.71} &       \underline{13.44} & 52.78 &    2.62 &       0.74 &       \underline{3.55} &        \underline{5.78} \\
     StarE  &   4 &   38.91 &    \underline{6.40} &       \underline{2.83} &       \underline{8.94} &       13.39 &  51.93 &    \textbf{5.06} &       \textbf{2.09} &             \textbf{7.34} &        \textbf{9.82} \\
    StarE &   6 &   38.20 &    6.87 &       3.46 &       8.98 &       \textbf{13.57} &   47.01 &    4.42 &       2.04 &       5.73 &        8.97 \\
     
    QBLP &   2 &   \underline{30.37} &    3.70 &       1.26 &       4.90 &        8.14 & 53.67 &    1.39 &       0.41 &       1.66 &        2.59\\
    QBLP  &   4 &   30.84 &    3.20 &       0.90 &         4.00 &        7.14 & \textbf{37.10} &    2.08 &       0.38 &            2.20 &        4.92 \\
    QBLP &   6 &   \textbf{26.34} &    4.34 &       1.66 &       5.53 &        9.25 &    \underline{39.12} &    1.95 &       0.41 &       2.15 &        4.10 \\
\bottomrule
\end{tabular}}
\caption{Results on the FI WD20K (66) V1 \& V2. \#QP denotes the number of qualifier pairs used in each statement (including padded pairs). Best results \textbf{in bold}, second best \underline{underlined}.}
\label{full_wd50_66_inductive}
\end{table*}

We implemented all approaches in Python building upon the open-source library \texttt{pykeen}~\cite{Ali*2021PyKEEN1.0Python} and make the code publicly available.\footnote{\url{https://github.com/mali-git/hyper_relational_ilp}} 
For each setting (i.e., dataset + number of qualifier pairs per triple), we performed a hyperparameter search using early stopping on the validation set and evaluated the final model on the test set.
We used AMR, MRR, and Hits@k as evaluation metrics, where the Adjusted Mean Rank (AMR)~\cite{berrendorf2020interpretable} is a recently proposed metric which sets the mean rank into relation with the expected mean rank of a random scoring model. 
Its value ranges from 0\%-200\%, and a lower value corresponds to better model performance.
Each model was trained at most 1000 epochs in the fully inductive setting, at most 600 epochs in the semi-inductive setting, and evaluated based on the early-stopping criterion with a frequency of 1, a patience of 200 epochs (in the semi-inductive setting, we performed all HPOs with a patience of 100 and 200 epochs), and a minimal improvement $\delta > 0.3\%$ optimizing the $hits@10$ metric.
For both inductive settings, we evaluated the effect of incorporating 0, 2, 4, and 6 qualifier pairs per triple. 

\subsection{Fully-Inductive Setting}\label{exp_full}

In the full inductive setting, we analyzed the effect of qualifiers for four different datasets (i.e., WD20K (100) V1 \& V2 and  WD20K (66) V1 \& V2, which have different ratios of qualifying statements and are of different sizes (see Section~\ref{sec:datasets}).
As triple-only baselines, we evaluated CompGCN~\cite{DBLP:conf/iclr/VashishthSNT20} and BLP~\cite{daza2020inductive}.
To evaluate the effect of qualifiers on the fully-inductive LP task, we evaluated StarE~\cite{galkin2020message} and QBLP.
It should be noted that StarE without the use of qualifiers is equivalent to CompGCN.

\textbf{General Overview}.
Tables~\ref{full_wd50_100_inductive}-\ref{full_wd50_66_inductive} show the results obtained for the four datasets.
The main findings are that (i) for all datasets, the use of qualifiers leads to increased performance, and (ii) the ratio of statements with qualifiers and the size of the dataset has a major impact on the performance.
CompGCN and StarE apply message-passing to obtain enriched entity representations while BLP and QBLP only apply a linear transformation.
Consequently, CompGCN and StarE require $\mathcal{S}_{\textit{inf}}$ to contain useful information in order to obtain the entity representations while BLP and QBLP are independent of $\mathcal{S}_{\textit{inf}}$.
In the following, we discuss the results for each dataset in detail.

\textbf{Results on WD20K (100) FI V1 \& V2}.
It can be observed that the performance gap between BLP/QBLP (0) and QBLP (2,4,6) is considerably larger than the gap between CompGCN and StarE.
This might be explained by the fact that QBLP does not take into account the graph structure provided by $\mathcal{S}_{\textit{inf}}$, therefore is heavily dependent on additional information, i.e. the qualifiers compensate for the missing graph information.
The overall performance decrease observable between V1 and V2 %
could be explained by the datasets' composition (Table~\ref{tab:datasets}), in particular, in the composition of the training and inference graphs:
$\mathcal{S}_{\textit{inf}}$ of V2 comprises more entities than V1, so that each test triple is ranked against more entities, i.e., the ranking becomes more difficult.
At the same time, the training graph of V1 is larger than that of V2, i.e., during training more entities (along their textual features) are seen which may improve generalization.

\textbf{Results on WD20K (66) FI V1 \& V2}.
Comparing StarE (2,4) to CompGCN (0), there is only a small improvement on this dataset. 
Also, the improvement of QBLP (2,4,6) compared to BLP and QBLP (0) is smaller than on the previous datasets.
This can be connected to the decreased ratio of statements with qualifiers. Besides, the training graph also has fewer qualifier pairs, $\mathcal{S}_{\textit{inf}}$ which is used by CompGCN and StarE 
for message passing consists of only 49\% of statements with at least one qualifier pair, and only 50\% of test statements have at least one qualifier pair which has an influence on all models.
This observation supports why StarE outperforms QBLP as the amount of provided qualifier statements cannot compensate for the graph structure in $\mathcal{S}_{\textit{inf}}$.

\begin{table*}[t]
\centering
\resizebox{\textwidth}{!}{
\begin{tabular}{lccccccccccc}
\toprule
\multirow{2}{*}{Model} & \multirow{2}{*}{\#QP} & \multicolumn{5}{c}{WD20K (33) SI } & \multicolumn{5}{c}{WD20K (25) SI} \\ \cmidrule(r){3-7}  \cmidrule(r){8-12}  
& & AMR(\%) &  MRR(\%) &  H@1(\%) &  H@5(\%) &  H@10(\%) & AMR(\%) &  MRR(\%) &  H@1(\%) &  H@5(\%) &  H@10(\%) \\
\midrule
  BLP &   0 &    \textbf{4.76} &   13.95 &       7.37 &      17.28 &       24.65 &  6.01 &   12.45 &       5.98 &     17.29 &       23.43 \\
  QBLP &   0 &    7.04 &   28.35 &      14.44 &      28.58 &       36.32 & 6.75 &   17.02 &       8.82 &      22.10 &       29.50 \\
 \midrule
  QBLP &   2 &   11.51 &   \textbf{35.95} &      \textbf{20.70} &      \textbf{34.98} &       \textbf{41.82} &    \underline{5.99} &   \underline{20.36} &      \underline{11.77} &      \textbf{24.86} &       \textbf{32.26} \\
  QBLP &   4 &   11.38 &   \underline{34.35} &      \underline{19.41} &      \underline{33.90} &       \underline{40.20}  & 12.18 &   \textbf{21.05} &      \textbf{12.32} &      24.07 &       30.09 \\
  QBLP &   6 &    \underline{4.98} &   25.94 &      15.20 &      30.06 &       38.70 & \textbf{5.73} &   19.50 &      11.14 &       \underline{24.73} &       \underline{31.60} \\
\bottomrule
\end{tabular}}
\caption{Results on the WD20K SI datasets. \#QP denotes the number of qualifier pairs used in each statement (including padded pairs).Best results \textbf{in bold}, second best \underline{underlined}.}
\label{full_wd50_100_inductive_s1}
\end{table*}

\subsection{Semi-inductive Setting}
\label{exp_semi}

In the semi-inductive setting, we evaluated BLP as a triple-only baseline and QBLP as a statement baseline (i.e., involving qualifiers) on the WD20K SI datasets. We did not evaluate CompGCN and StarE since message-passing-based approaches are not directly applicable in the absence of $\mathcal{S}_{\textit{inf}}$.
The results highlight that aggregating qualifier information improves the prediction of semi-inductive links despite the fact that the ratio of statements with qualifiers is not very large (37\% for SI WD20K (33), and 30\% for SI WD20K (25)).
In the case of SI WD20K (33), the baselines are outperformed even by a large margin.
Overall, the results might indicate that in semi-inductive settings, performance improvements can already be obtained with a decent amount of statements with qualifiers.

\begin{figure}
    \centering
    \includegraphics[width=.5\textwidth]{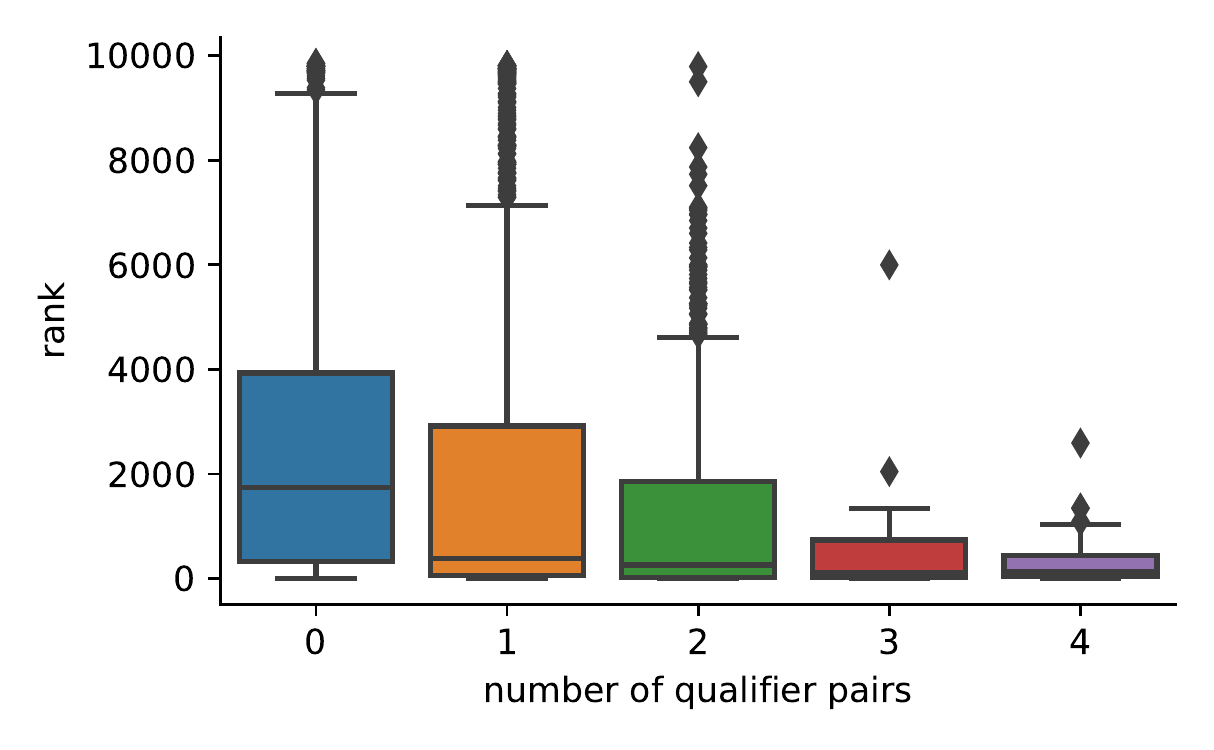}
    \caption{
Distribution of individual ranks for head/tail prediction with  StarE on WD20K (66) V2.
The statements are grouped by the number of qualifier pairs.
}
    \label{fig:qualfier_rank}
\end{figure}

\begin{figure}[t]
    \centering
    \includegraphics[width=.5\linewidth]{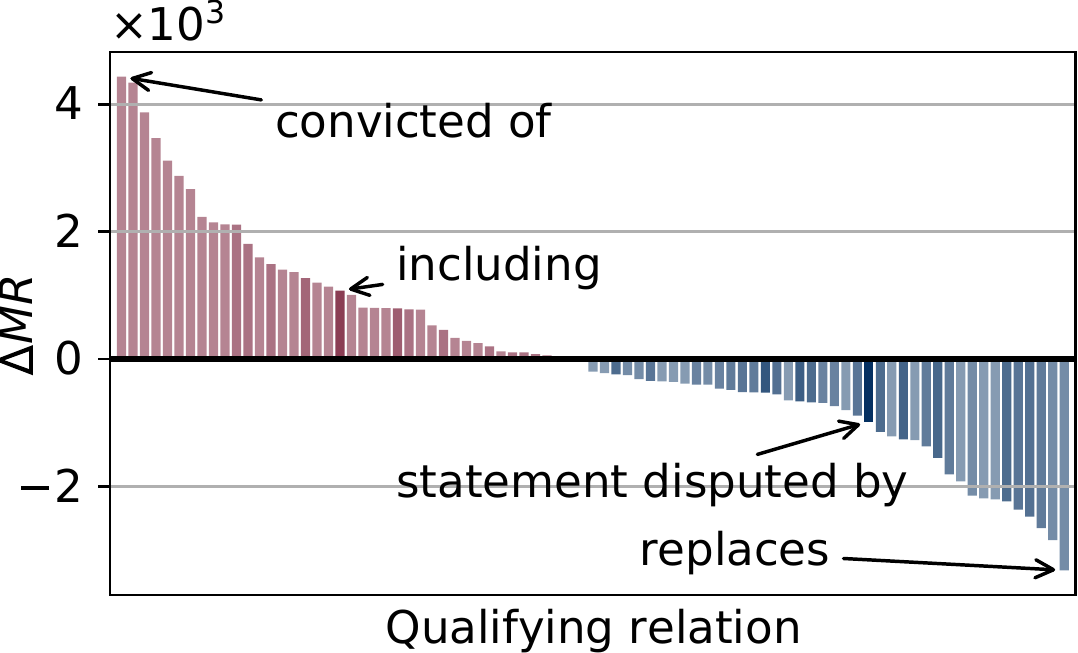}
    \caption{
    Rank deviation when masking qualifier pairs containing a certain relation.
    Transparency is proportional to the occurrence frequency, bar height/color indicates difference in MR \emph{for evaluation statements using this qualifying relation} if the pair is masked. More negative deltas correspond to better predictions.
    }
    \label{fig:qualifier_relation_importance_restricted}
\end{figure}

\subsection{Qualitative Analysis}

We obtain deeper insights on the impact of qualifiers by analyzing the StarE model on the fully-inductive WD20K (66) V2 dataset.
In particular, we study individual ranks for head/tail prediction of statements with and without qualifiers (cf. Fig.~\ref{fig:qualfier_rank}) varying the model from zero to four pairs.
First, we group the test statements by the number of available qualifier pairs.
We observe generally smaller ranks which, in turn, correspond to better predictions when more qualifier pairs are available.
In particular, just one qualifier pair is enough to significantly reduce the individual ranks. %
Note that we have less statements with many qualifiers, cf.  Appendix~\ref{app:datas}.

We then study how particular qualifiers affect ranking and predictions. 
For that, we measure ranks of predictions for distinct statements in the \emph{test set} with and without masking the qualifier relation from the inference graph $\mathcal{S}_{\mathit{inf}}$.
We then compute $\Delta \textit{MR}$ and group them by used qualifier relations (Fig.~\ref{fig:qualifier_relation_importance_restricted}).
Interestingly, certain qualifiers, e.g., \texttt{convicted of} or \texttt{including}, deteriorate the performance which we attribute to the usage of rare, qualifier-only entities. 
Conversely, having qualifiers like \texttt{replaces} reduces the rank by about 4000 which greatly improves prediction accuracy. 
We hypothesize it is an effect of qualifier entities: helpful qualifiers employ well-connected nodes in the graph which benefit from message passing.

\begin{table}%
    \centering
    \small
    \begin{tabular}{llr}
    \toprule
    \multicolumn{3}{c}{WD20K (100) V1 FI}\\
    \midrule
Wikidata ID & relation name &   $\Delta$MR \\
\midrule
         P2868 &        subject has role &   0.12 \\
          P463 &               member of &  -0.04 \\
         P1552 &             has quality &  -0.34 \\
          \midrule
         P2241 &  reason for deprecation & -26.44 \\
           P47 &      shares border with & -28.91 \\
          P750 &          distributed by & -29.12 \\
     \midrule
    \multicolumn{3}{c}{WD20K (66) V2 FI}\\
    \midrule
          P805 &  statement is subject of &  13.11 \\
         P1012 &                including &   5.95 \\
          P812 &           academic major &   5.07 \\
         \midrule
           P17 &                  country & -19.96 \\
         P1310 &    statement disputed by & -20.92 \\
         P1686 &                 for work & -56.87 \\
     \bottomrule
    \end{tabular}
    \caption{
    Top 3 worst and best qualifier relations affecting the overall mean rank (the last column). Negative $\Delta$MR with larger absolute value correspond to better predictions.
    }
    \label{tab:qualifier_relation_importance_overall}
\end{table}

Finally, we study the average impact of qualifiers on the whole graph, i.e.,  we take the whole \emph{inference graph} and mask out all qualifier pairs containing one relation and compare the overall evaluation result on the test set (in contrast to Fig.~\ref{fig:qualifier_relation_importance_restricted}, we count ranks of all test statements, not only those which have that particular qualifier) against the non-masked version of the same graph. 
We then sort relations by $\Delta \textit{MR}$ and find top 3 most confusing and most helpful relations across two datasets (cf. Table~\ref{tab:qualifier_relation_importance_overall}). 
On the smaller WD20K (100) V1 where all statements have at least one qualifier pair, most relations tend to improve MR. 
For instance, qualifiers with the \texttt{distributed by} relations reduce MR by about 29 points.
On the larger WD20K (66) V2 some qualifier relations, e.g., \texttt{statement is subject of}, tend to introduce more noise and worsen MR which we attribute to the increased sparsity of the graph given an already rare qualifier entity.
That is, such rare entities might not benefit enough from message passing. 

\section{Related Work}

We focus on semi- and fully inductive link prediction approaches and disregard classical approaches that are fully transductive, which have been extensively studied in the literature~\cite{DBLP:journals/corr/abs-2006-13365,DBLP:journals/corr/abs-2002-00388}.

In the domain of triple-only KGs, both settings have recently received a certain traction.
One of the main challenges for realistic KG embedding is the impossibility of learning representations of unseen entities since they are not present in the train set.

In the semi-inductive setting, several methods alleviating the issue were proposed. 
When a new node arrives with a certain set of edges to known nodes, \cite{DBLP:conf/emnlp/AlbooyehGK20} enhanced the training procedure such that an embedding of an unseen node is a linear aggregation of neighbouring nodes. 
If there is no connection to the seen nodes, \cite{DBLP:journals/corr/abs-2009-09263} propose to \emph{densify} the graph with additional edges obtained from pairwise similarities of node features.
Another approach applies a special meta-learning framework~\cite{DBLP:conf/nips/BaekLH20} when during training a meta-model has to learn representations decoupled from concrete training  entities but transferable to unseen entities.
Finally, reinforcement learning methods~\cite{DBLP:conf/semweb/BhowmikM20} were employed to learn relation paths between seen and unseen entities. 

In the fully inductive setup, the evaluation graph is a separate subgraph disjoint with the training one, which makes trained entity embeddings even less useful. 
In such cases, the majority of existing methods~\cite{yao2019kgbert,clouatre2020mlmlm,daza2020inductive,DBLP:conf/emnlp/ZhangLZ00H20} resort to pre-trained language models (LMs) (e.g., BERT~\cite{DBLP:conf/naacl/DevlinCLT19}) as \emph{universal featurizers}. 
That is, textual entity descriptions (often available in KGs at least in English) are passed through an LM to obtain initial semantic node features.
Nevertheless, mining and employing structural graph features, e.g., shortest paths within sampled subgraphs, has been shown~\cite{DBLP:conf/icml/TeruDH20} to be beneficial as well.
This work is independent from the origin of node features and is able to leverage both, although the new datasets employ Sentence BERT~\cite{reimers-2019-sentence-bert} for featurizing.

All the described approaches operate on triple-based KGs whereas our work studies inductive LP problems on enriched, hyper-relational KGs where we show that incorporating such hyper-relational information indeed leads to better performance.

\section{Conclusion}

In this work, we presented a study of the inductive link prediction problem over hyper-relational KGs.
In particular, we proposed a theoretical framework to categorize various LP tasks to alleviate an existing terminology discrepancy pivoting on two settings, namely, semi- and fully-inductive LP.
Then, we designed WD20K, a collection of hyper-relational benchmarks based on Wikidata for inductive LP with a diverse set of parameters and complexity.
Probing statement-aware models against triple-only baselines, we demonstrated that hyper-relational facts indeed improve LP performance in both inductive settings by a considerable margin.
Moreover, our qualitative analysis showed that the achieved gains are consistent across different setups and still interpretable.

Our findings open up interesting prospects for employing inductive LP and hyper-relational KGs along several axes, e.g., large-scale KGs of billions statements, new application domains including life sciences, drug discovery, and KG-based NLP applications like question answering or entity linking.

In the future, we plan to extend inductive LP to consider unseen relations and qualifiers; tackle the problem of suggesting best qualifiers for a statement; and provide more solid theoretical foundations of representation learning over hyper-relational KGs.

\section*{Acknowledgements}
This work was funded by the German Federal Ministry of Education and Research (BMBF) under Grant No. 01IS18036A and Grant No. 01IS18050D (project “MLWin”).
The authors of this work take full responsibilities for its content.

\bibliography{main}

\begin{thebibliography}{10}
\providecommand{\url}[1]{\texttt{#1}}
\providecommand{\urlprefix}{URL }
\providecommand{\doi}[1]{https://doi.org/#1}

\bibitem{DBLP:conf/emnlp/AlbooyehGK20}
Albooyeh, M., Goel, R., Kazemi, S.M.: Out-of-sample representation learning for
  knowledge graphs. In: Cohn, T., He, Y., Liu, Y. (eds.) Proceedings of the
  2020 Conference on Empirical Methods in Natural Language Processing:
  Findings, {EMNLP} 2020, Online Event, 16-20 November 2020. pp. 2657--2666.
  Association for Computational Linguistics (2020)

\bibitem{DBLP:journals/corr/abs-2006-13365}
Ali, M., Berrendorf, M., Hoyt, C.T., Vermue, L., Galkin, M., Sharifzadeh, S.,
  Fischer, A., Tresp, V., Lehmann, J.: Bringing light into the dark: {A}
  large-scale evaluation of knowledge graph embedding models under a unified
  framework. CoRR  \textbf{abs/2006.13365} (2020)

\bibitem{Ali*2021PyKEEN1.0Python}
Ali*, M., Berrendorf*, M., Hoyt*, C.T., Vermue*, L., Sharifzadeh, S., Tresp,
  V., Lehmann, J.: Pykeen 1.0: A python library for training and evaluating
  knowledge graph embeddings. Journal of Machine Learning Research
  \textbf{22}(82),  1–6 (2021), \url{http://jmlr.org/papers/v22/20-825.html},
  * equal contribution

\bibitem{DBLP:conf/nips/BaekLH20}
Baek, J., Lee, D.B., Hwang, S.J.: Learning to extrapolate knowledge:
  Transductive few-shot out-of-graph link prediction. In: Larochelle, H.,
  Ranzato, M., Hadsell, R., Balcan, M., Lin, H. (eds.) Advances in Neural
  Information Processing Systems 33: Annual Conference on Neural Information
  Processing Systems 2020, NeurIPS 2020, December 6-12, 2020, virtual (2020)

\bibitem{dti-overview}
Bagherian, M., Sabeti, E., Wang, K., Sartor, M.A., Nikolovska-Coleska, Z.,
  Najarian, K.: {Machine learning approaches and databases for prediction of
  drug–target interaction: a survey paper}. Briefings in Bioinformatics
  \textbf{22}(1),  247--269 (01 2020). \doi{10.1093/bib/bbz157},
  \url{https://doi.org/10.1093/bib/bbz157}

\bibitem{DBLP:conf/nips/BelkinN01}
Belkin, M., Niyogi, P.: Laplacian eigenmaps and spectral techniques for
  embedding and clustering. In: Dietterich, T.G., Becker, S., Ghahramani, Z.
  (eds.) Advances in Neural Information Processing Systems 14 [Neural
  Information Processing Systems: Natural and Synthetic, {NIPS} 2001, December
  3-8, 2001, Vancouver, British Columbia, Canada]. pp. 585--591. {MIT} Press
  (2001)

\bibitem{berrendorf2020interpretable}
Berrendorf, M., Faerman, E., Vermue, L., Tresp, V.: Interpretable and fair
  comparison of link prediction or entity alignment methods with adjusted mean
  rank. In: 2020 IEEE/WIC/ACM International Joint Conference on Web
  Intelligence and Intelligent Agent Technology (WI-IAT'20). IEEE (2020)

\bibitem{DBLP:conf/semweb/BhowmikM20}
Bhowmik, R., de~Melo, G.: Explainable link prediction for emerging entities in
  knowledge graphs. In: Pan, J.Z., Tamma, V.A.M., d'Amato, C., Janowicz, K.,
  Fu, B., Polleres, A., Seneviratne, O., Kagal, L. (eds.) The Semantic Web -
  {ISWC} 2020 - 19th International Semantic Web Conference. Lecture Notes in
  Computer Science, vol. 12506, pp. 39--55. Springer (2020)

\bibitem{DBLP:conf/nips/BordesUGWY13}
Bordes, A., Usunier, N., Garc{\'{\i}}a{-}Dur{\'{a}}n, A., Weston, J.,
  Yakhnenko, O.: Translating embeddings for modeling multi-relational data. In:
  Burges, C.J.C., Bottou, L., Ghahramani, Z., Weinberger, K.Q. (eds.) Advances
  in Neural Information Processing Systems 26: 27th Annual Conference on Neural
  Information Processing Systems 2013. Proceedings of a meeting held December
  5-8, 2013, Lake Tahoe, Nevada, United States. pp. 2787--2795 (2013)

\bibitem{DBLP:journals/corr/abs-2006-09252}
Bouritsas, G., Frasca, F., Zafeiriou, S., Bronstein, M.M.: Improving graph
  neural network expressivity via subgraph isomorphism counting. CoRR
  \textbf{abs/2006.09252} (2020)

\bibitem{DBLP:journals/corr/abs-2005-03675}
Chami, I., Abu{-}El{-}Haija, S., Perozzi, B., R{\'{e}}, C., Murphy, K.: Machine
  learning on graphs: {A} model and comprehensive taxonomy. CoRR
  \textbf{abs/2005.03675} (2020)

\bibitem{clouatre2020mlmlm}
Clouatre, L., Trempe, P., Zouaq, A., Chandar, S.: Mlmlm: Link prediction with
  mean likelihood masked language model (2020)

\bibitem{daza2020inductive}
Daza, D., Cochez, M., Groth, P.: Inductive entity representations from text via
  link prediction (2020)

\bibitem{DBLP:conf/aaai/DettmersMS018}
Dettmers, T., Minervini, P., Stenetorp, P., Riedel, S.: Convolutional 2d
  knowledge graph embeddings. In: {AAAI}. pp. 1811--1818. {AAAI} Press (2018)

\bibitem{DBLP:conf/naacl/DevlinCLT19}
Devlin, J., Chang, M., Lee, K., Toutanova, K.: {BERT:} pre-training of deep
  bidirectional transformers for language understanding. In: Burstein, J.,
  Doran, C., Solorio, T. (eds.) Proceedings of the 2019 Conference of the North
  American Chapter of the Association for Computational Linguistics: Human
  Language Technologies, {NAACL-HLT} 2019, Minneapolis, MN, USA, June 2-7,
  2019, Volume 1 (Long and Short Papers). pp. 4171--4186. Association for
  Computational Linguistics (2019)

\bibitem{DBLP:conf/kdd/0001GHHLMSSZ14}
Dong, X., Gabrilovich, E., Heitz, G., Horn, W., Lao, N., Murphy, K., Strohmann,
  T., Sun, S., Zhang, W.: Knowledge vault: a web-scale approach to
  probabilistic knowledge fusion. In: Macskassy, S.A., Perlich, C., Leskovec,
  J., Wang, W., Ghani, R. (eds.) The 20th {ACM} {SIGKDD} International
  Conference on Knowledge Discovery and Data Mining, {KDD} '14, New York, NY,
  {USA} - August 24 - 27, 2014. pp. 601--610. {ACM} (2014)

\bibitem{galkin2020message}
Galkin, M., Trivedi, P., Maheshwari, G., Usbeck, R., Lehmann, J.: Message
  passing for hyper-relational knowledge graphs. In: Webber, B., Cohn, T., He,
  Y., Liu, Y. (eds.) Proceedings of the 2020 Conference on Empirical Methods in
  Natural Language Processing, {EMNLP} 2020, Online, November 16-20, 2020. pp.
  7346--7359. Association for Computational Linguistics (2020)

\bibitem{DBLP:journals/corr/abs-2012-05716}
Gaudelet, T., Day, B., Jamasb, A.R., Soman, J., Regep, C., Liu, G., Hayter,
  J.B.R., Vickers, R., Roberts, C., Tang, J., Roblin, D., Blundell, T.L.,
  Bronstein, M.M., Taylor{-}King, J.P.: Utilising graph machine learning within
  drug discovery and development. CoRR  \textbf{abs/2012.05716} (2020)

\bibitem{DBLP:conf/icml/GilmerSRVD17}
Gilmer, J., Schoenholz, S.S., Riley, P.F., Vinyals, O., Dahl, G.E.: Neural
  message passing for quantum chemistry. In: Precup, D., Teh, Y.W. (eds.)
  Proceedings of the 34th International Conference on Machine Learning, {ICML}
  2017, Sydney, NSW, Australia, 6-11 August 2017. Proceedings of Machine
  Learning Research, vol.~70, pp. 1263--1272. {PMLR} (2017)

\bibitem{DBLP:journals/corr/abs-2002-00388}
Ji, S., Pan, S., Cambria, E., Marttinen, P., Yu, P.S.: A survey on knowledge
  graphs: Representation, acquisition and applications. CoRR
  \textbf{abs/2002.00388} (2020)

\bibitem{DBLP:journals/corr/abs-1907-11692}
Liu, Y., Ott, M., Goyal, N., Du, J., Joshi, M., Chen, D., Levy, O., Lewis, M.,
  Zettlemoyer, L., Stoyanov, V.: Roberta: {A} robustly optimized {BERT}
  pretraining approach. CoRR  \textbf{abs/1907.11692} (2019),
  \url{http://arxiv.org/abs/1907.11692}

\bibitem{DBLP:conf/icml/NickelTK11}
Nickel, M., Tresp, V., Kriegel, H.: A three-way model for collective learning
  on multi-relational data. In: Getoor, L., Scheffer, T. (eds.) Proceedings of
  the 28th International Conference on Machine Learning, {ICML} 2011, Bellevue,
  Washington, USA, June 28 - July 2, 2011. pp. 809--816. Omnipress (2011)

\bibitem{reimers-2019-sentence-bert}
Reimers, N., Gurevych, I.: Sentence-bert: Sentence embeddings using siamese
  bert-networks. In: Proceedings of the 2019 Conference on Empirical Methods in
  Natural Language Processing. Association for Computational Linguistics (11
  2019), \url{https://arxiv.org/abs/1908.10084}

\bibitem{DBLP:conf/icml/TeruDH20}
Teru, K., Denis, E., Hamilton, W.: Inductive relation prediction by subgraph
  reasoning. In: Proceedings of the 37th International Conference on Machine
  Learning, {ICML} 2020, 13-18 July 2020, Virtual Event. Proceedings of Machine
  Learning Research, vol.~119, pp. 9448--9457. {PMLR} (2020)

\bibitem{DBLP:conf/iclr/VashishthSNT20}
Vashishth, S., Sanyal, S., Nitin, V., Talukdar, P.P.: Composition-based
  multi-relational graph convolutional networks. In: 8th International
  Conference on Learning Representations, {ICLR} 2020, Addis Ababa, Ethiopia,
  April 26-30, 2020. OpenReview.net (2020),
  \url{https://openreview.net/forum?id=BylA\_C4tPr}

\bibitem{DBLP:journals/cacm/VrandecicK14}
Vrandecic, D., Kr{\"{o}}tzsch, M.: Wikidata: a free collaborative
  knowledgebase. Commun. {ACM}  \textbf{57}(10),  78--85 (2014)

\bibitem{DBLP:journals/corr/abs-2009-09263}
Wang, B., Wang, G., Huang, J., You, J., Leskovec, J., Kuo, C.J.: Inductive
  learning on commonsense knowledge graph completion. CoRR
  \textbf{abs/2009.09263} (2020)

\bibitem{yao2019kgbert}
Yao, L., Mao, C., Luo, Y.: Kg-bert: Bert for knowledge graph completion (2019)

\bibitem{DBLP:conf/emnlp/ZhangLZ00H20}
Zhang, Z., Liu, X., Zhang, Y., Su, Q., Sun, X., He, B.: Pretrain-kge: Learning
  knowledge representation from pretrained language models. In: Cohn, T., He,
  Y., Liu, Y. (eds.) Proceedings of the 2020 Conference on Empirical Methods in
  Natural Language Processing: Findings, {EMNLP} 2020, Online Event, 16-20
  November 2020. pp. 259--266. Association for Computational Linguistics (2020)

\end{thebibliography}
\bibliographystyle{splncs04}
\clearpage

\appendix

\section{Training}
\label{app:training}
In the sLCWA, negative training examples are created for each true fact $(h,r,t)\in KG$ by corrupting the head or tail entity resulting in the triples $(h',r,t)/(h,r,t')$.
In the LCWA, for each triple $(h,r,t)\in KG$ all triples $(h,r,t')\notin KG$ are considered as non-existing, i.e., as negative examples.

Under the sLCWA, we trained the models using the margin ranking loss~\cite{DBLP:conf/nips/BordesUGWY13}:
\begin{equation}
L(f(t_i^{+}),f(t_i^{-})) = \max(0, \lambda + f(t_i^{-}) - f(t_i^{+})) \enspace,
\end{equation}

where $f(t_i^{+})$ denotes the model's score for a positive training example and $f(t_i^{-})$ for a negative one.

For training under the LCWA, we used the binary cross entropy loss~\cite{DBLP:conf/aaai/DettmersMS018}:

\begin{equation}
\begin{aligned}
L(f(t_{i}), l_i) = 
&-(l_{i} \cdot \log(\sigma(f(t_{i}))) \\
&+ (1 - l_{i}) \cdot \log(1 - \sigma(f(t_{i})))),
\end{aligned}
\end{equation}

where $l_i$ corresponds to the label of the triple $t_i$.

\section{Hyperparameter Ranges}

The following tables summarizes the hyper-parameter ranges explored during hyper-parameter optimization. The best hyper-parameters for each of our 46 ablation studies will be available online upon publishing.

\begin{table}[t]
    \centering
    \begin{tabular}{lr}
\toprule
Hyper-Parameter &  Value \\
\midrule
  GCN layers & \{2,3\} \\
  Embedding dim. & \{32, 64, ... , 256 \} \\
  Transformer hid. dim. & \{512, 576, ... , 1024  \} \\
  Num. attention heads & \{2, 4\} \\
  Num. transformer heads & \{2, 4\} \\
  Num. transformer layers & \{2, 3, 4\} \\
  Qualifier aggr. & \{sum, attention\} \\
  Qualifier weight & 0.8 \\
  Dropout & \{0.1, 0.2, ... , 0.5 \} \\
  Attention slope & \{0.1, 0.2, 0.3, 0.4 \} \\
  Training approaches & \{sLCWA, LCWA\} \\
  Loss fcts. & \{MRL, BCEL\} \\
  Learning rate (log scale) & [0.0001, 1.0) \\
  Label smoothing & \{0.1, 0.15\} \\
  Batch size & \{128, 192, ... , 1024\} \\
  Max Epochs FI setting & 1000 \\
  Max Epochs SI setting & 600 \\
\bottomrule
\end{tabular}
    \caption{Hyperparameter ranges explored during hyper-parameter optimization. FI denotes the fully-inductive setting and SI the semi-inductive setting. For the sLCWA training approach, we trained the models with the margin ranking loss (MRL), and with the LCWA we used the BCEL (Binary Cross Entropy loss}.
    \label{tab:my_label}
\end{table}

\section{Infrastructure and Parameters}

We train each model on machines running Ubuntu 18.04 equipped with a GeForce RTX 2080 Ti with 12GB RAM.
In total, we performed 46 individual hyperparameter optimizations (one for each dataset / model / number-of-qualifier combination).
Depending on the exact configuration, the individual models have between 500k and 5M parameters and take up to 2 hours for training.

\section{Qualifier Ratio}
\label{app:datas}

\begin{figure*}
    \centering
    \includegraphics[width=\linewidth]{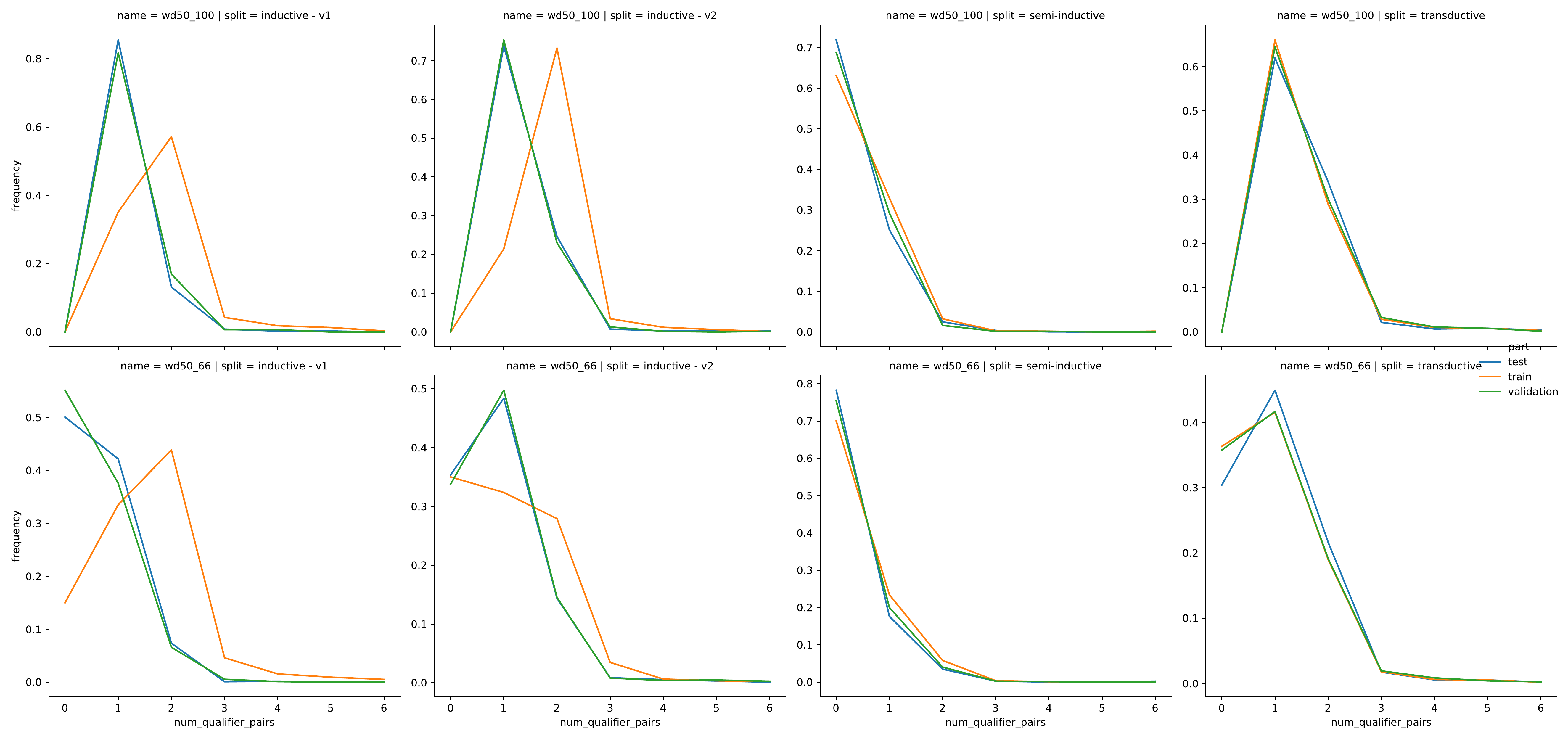}
    \caption{
    Percentage of statements with the given number of available qualifier pairs for all datasets and splits.
    }
    \label{fig:qualifier_ratio_all_datasets}
\end{figure*}
Figure~\ref{fig:qualifier_ratio_all_datasets} shows the ratio of statements with a given number of available qualifier pairs for all datasets and splits.
We generally observe that there are only few statements with a large number of qualifier pairs, while most of them have zero to two qualifier pairs.

\end{document}